\documentclass[10pt,twocolumn,letterpaper]{article}

\usepackage{cite}
\usepackage{changepage}

\usepackage{times}
\usepackage{epsfig}
\usepackage{graphicx}
\usepackage{amsmath}
\usepackage{amssymb}
\usepackage{color}

\usepackage[table, dvipsnames]{xcolor}
\usepackage{arydshln}

\usepackage{comment}
\usepackage{amsmath,amssymb} 
\usepackage{multirow, array, booktabs,xspace,bm}
\usepackage{pifont}
\usepackage{amsfonts}
\usepackage{makecell}
\usepackage{dcolumn}
\usepackage{eqparbox}
\usepackage{subcaption}
\usepackage{enumitem}

\usepackage{algorithm,algpseudocode}

\newcommand{\cmark}{\ding{51}}%
\newcommand{\xmark}{\ding{55}}%
\newcolumntype{P}[1]{>{\centering\arraybackslash}p{#1}}
\newcolumntype{M}[1]{>{\centering\arraybackslash}m{#1}}
\newcolumntype{L}[1]{>{\raggedright\arraybackslash}p{#1}}

\definecolor{red2}{rgb}{1.0,0.20,0.20}

\definecolor{blue2}{rgb}{0.220,0.407,0.82}
\definecolor{lightgray}{rgb}{0.935, 0.935, 0.935}

\definecolor{cellcol}{gray}{.92}
\newcommand{\graycell}[1]{\cellcolor{cellcol}{#1}}

\usepackage[pagenumbers]{cvpr} 

\usepackage[dvipsnames]{xcolor}

\definecolor{cvprblue}{rgb}{0.21,0.49,0.74}
\usepackage[pagebackref,breaklinks,colorlinks,citecolor=cvprblue]{hyperref}

\def\paperID{7156} 
\def\confName{CVPR}
\def\confYear{2024}

\title{Modeling Multimodal Social Interactions: New Challenges and Baselines \\with Densely Aligned Representations}

\author{Sangmin Lee$^{1}$ \;\; Bolin Lai$^{2}$ \;\; Fiona Ryan$^{2}$ \;\; Bikram Boote$^{1}$ \;\; James M. Rehg$^{1}$\\ 
$^{1}$University of Illinois Urbana-Champaign  \ $^{2}$Georgia Institute of Technology\\
{\tt\small\{sangminl,boote,jrehg\}@illinois.edu \ \{bolin.lai,fkryan\}@gatech.edu}}

\begin{document}
\maketitle

\begin{abstract}
Understanding social interactions involving both verbal and non-verbal cues is essential for effectively interpreting social situations. However, most prior works on multimodal social cues focus predominantly on single-person behaviors or rely on holistic visual representations that are not aligned to utterances in multi-party environments. Consequently, they are limited in modeling the intricate dynamics of multi-party interactions. In this paper, we introduce three new challenging tasks to model the fine-grained dynamics between multiple people: speaking target identification, pronoun coreference resolution, and mentioned player prediction. We contribute extensive data annotations to curate these new challenges in social deduction game settings. Furthermore, we propose a novel multimodal baseline that leverages densely aligned language-visual representations by synchronizing visual features with their corresponding utterances. This facilitates concurrently capturing verbal and non-verbal cues pertinent to social reasoning. Experiments demonstrate the effectiveness of the proposed approach with densely aligned multimodal representations in modeling fine-grained social interactions. Project website: \href{https://sangmin-git.github.io/projects/MMSI}{\color{magenta}{https://sangmin-git.github.io/projects/MMSI}}.
\end{abstract}

\section{Introduction}
Real-world social interactions involve intricate behaviors between multiple people. People communicate not only through verbal cues (\textit{e.g.}, language) but also through non-verbal cues (\textit{e.g.}, gesture, gaze). While spoken language conveys explicit meaning, inferring the full social context from language alone can sometimes be ambiguous. Non-verbal cues can often play a crucial role in clarifying these subtle social nuances and providing additional context. Consequently, comprehensively understanding social interactions involving multimodal social cues is essential to interpret social situations appropriately.

\begin{figure}[t!]
\begin{minipage}[b]{1.0\linewidth}
\centering
\centerline{\includegraphics[width=8.5cm]{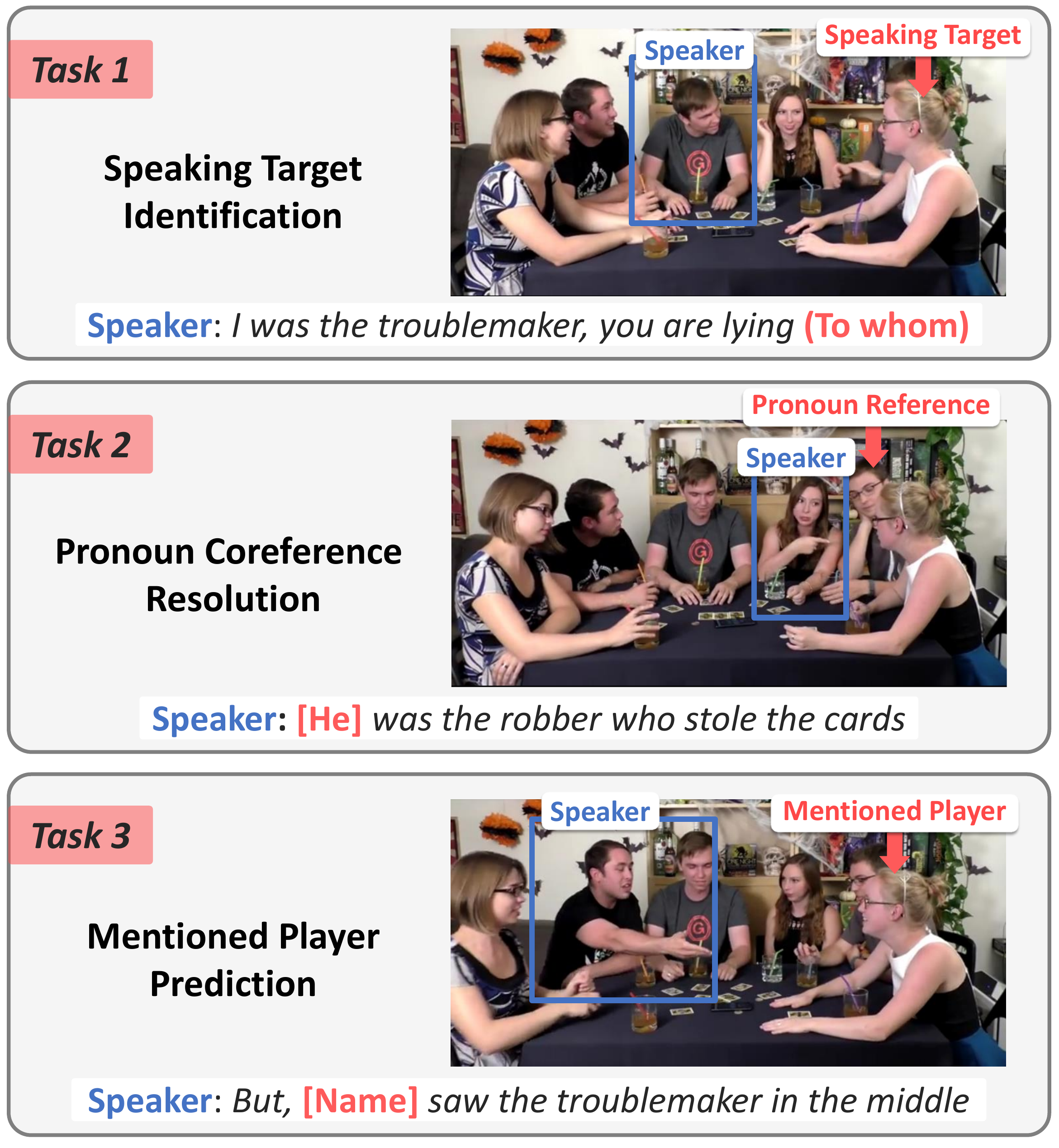}}
\end{minipage}
\vspace{-0.6cm}
\caption{Concepts of the proposed three social tasks in multi-party environments: speaking target identification, pronoun coreference resolution, and mentioned player prediction.}
\label{figure_task}
\vspace{-0.25cm}
\end{figure}

There have been attempts to investigate social behaviors in multimodal aspects by jointly considering language and visual cues. Some works tried to learn the relationships between spoken language and visual gestures for gesture generation \cite{ahuja2020no, liu2022learning,ahuja2023continual} and gesture-language grounding \cite{lee2021crossmodal}. Other multimodal approaches utilized the interconnection between spoken language and visual expressions to recognize human sentiment and emotion \cite{zadeh2018multimodal, saha2020towards, hu2022unimse, paraskevopoulos2022mmlatch, hu2022mm, chen2023multivariate}. However, these works mainly focus on single-person behaviors or rely on holistic visual representations in multi-party settings, rather than modeling the fine-grained dynamics of social interactions among multiple people. Modeling such intricate multi-party dynamics requires understanding the relationships between language and visual cues at an individual level, rather than using global representations.

Recently, a multimodal work \cite{lai2023werewolf} addressed social behaviors in social deduction games, which provide an effective testbed for studying multimodal social interactions. In these games, players take on roles and attempt to deduce the roles of their opponents by engaging in communication, deception, inference, and collaboration. These games encompass rich social interactions including verbal and non-verbal cues in multi-party settings. Lai \textit{et al.} \cite{lai2023werewolf} leveraged language and visual cues to predict persuasion strategies at the utterance level. However, their work has limitations in modeling multi-party interactions in terms of its task and methodology. Although persuasion strategies emerge in communication, the task primarily focuses on understanding the social behaviors of a single person rather than the dynamics among people. Moreover, their approach is limited in distinguishing and recognizing fine-grained interactions because it utilizes holistic visual representations for the entire scene, despite the presence of multiple people.

\begin{figure}[t!]
\begin{minipage}[b]{1.0\linewidth}
\centering
\centerline{\includegraphics[width=8.0cm]{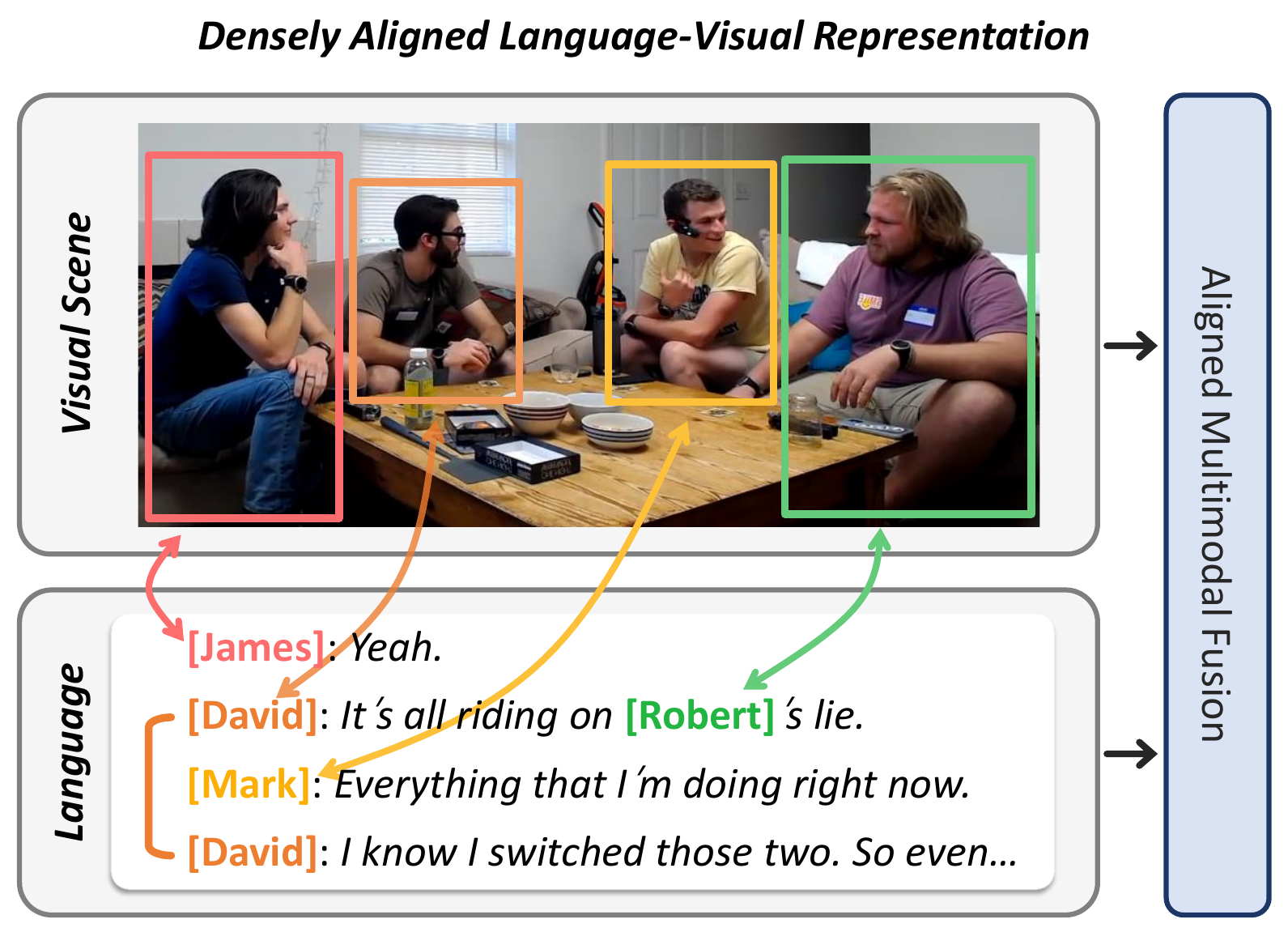}}
\end{minipage}
\vspace{-0.6cm}
\caption{Concept of densely aligned language-visual representations. People are matched in the language and visual domains.}
\label{figure_dense}
\vspace{-0.25cm}
\end{figure}

To address these issues, we introduce three new multimodal tasks that encompass the dynamics of multi-party interactions, along with a novel baseline model. We extend the social deduction game datasets \cite{lai2023werewolf} with extensive data annotations to curate new social tasks focused on identifying referents in multi-party conversations. Appropriately identifying referents is important because it significantly affects interpreting the social intent and context of conversations. Figure \ref{figure_task} shows the overall concepts of our curated social tasks. The three tasks are as follows. 
\vspace{0.05cm}
\begin{enumerate}[label=\textbf{\arabic*.}]
    \item \textbf{\textit{Speaking target identification}:}\ Identifying who a speaker is talking to in a conversation.
    \item \textbf{\textit{Pronoun coreference resolution}:}\ Determining who a pronoun refers to in a conversation.
    \item \textbf{\textit{Mentioned player prediction}:}\ Predicting who is mentioned by name in a conversation. 
\end{enumerate}
\vspace{0.05cm}

\noindent These tasks are challenging as they require understanding the fine-grained dynamics of player interactions. For instance, when an utterance is accompanied by visual cues like pointing gestures, it is necessary to comprehensively interpret the utterance content and the speaker's gestures for holistic reasoning. To this end, we need to figure out who the speaker is and who is being pointed at within the visual scene based on the utterance context. This involves matching utterances with the visually identified individuals. Thus, it is required to align language references with player visuals and to derive densely aligned multimodal representations from such aligned features. Figure \ref{figure_dense} shows the concept of densely aligned language-visual representations.

To this end, we propose a novel baseline model leveraging densely aligned language-visual representations. We detect and visually track each player in the video to distinguish individual players. By initially aligning player visual positions with their language references, we continuously track player visuals in sync with the dialogue. This alignment allows the model to visually identify the speaker and the other players (\textit{i.e.}, listeners) for given utterances. Encoding speaker gestures and the relative positions of the listeners enables deciphering visual relationships for understanding non-verbal dynamics. We then combine this with the linguistic context of the conversation to obtain densely aligned representations. Consequently, we can effectively predict referents by leveraging densely aligned representations containing both verbal and non-verbal dynamics.

\noindent The major contributions of this paper are as follows.
\vspace{0.05cm}
\begin{itemize}
	\item We introduce new social tasks in multi-party settings via extensive data annotations: \textit{speaking target identification}, \textit{pronoun coreference resolution}, and \textit{mentioned player prediction}. These tasks are challenging as they require understanding the fine-grained dynamics of interactions.
	\item	We propose a novel multimodal baseline model leveraging language and visual cues for understanding multi-party social interactions. To the best of our knowledge, this is the first work to address the multimodal dense alignment between language and visual social cues.
\end{itemize}

\begin{table*}[t]{
\renewcommand{\arraystretch}{1.2}
\renewcommand{\tabcolsep}{2.5mm}
\centering
\resizebox{0.98\linewidth}{!}{
\begin{tabular}{c  c  c  c  c c}
\toprule
\multirow{2}{*}{\bf Annotation Type} & \multirow{2}{*}{\bf Utterance Example} & \multicolumn{2}{c}{\bf \makecell{YouTube DB}}  & \multicolumn{2}{c}{\bf \makecell{Ego4D DB}}\\ 
\cmidrule(lr){3-4}
\cmidrule(lr){5-6}
& & \bf Count & \bf Krippendorff's $\boldsymbol{\alpha}$ &\bf Count & \bf Krippendorff's $\boldsymbol{\alpha}$ \\ \bottomrule

\makecell{Speaking Target \\ Identification} & \makecell{\textit{Why are you helping the Werewolves} \\ \textit{out?} (To \textbf{\textcolor{red2}{[Name]}})}   & 3,255 &  0.922 & 832 & 0.907 \\ \hline

\makecell{Pronoun Coreference \\ Resolution} & \makecell{\textit{I'm a Villager which makes me think} \\ \textit{\textbf{ \underline{he}}} ($\rightarrow$ \textbf{\textcolor{red2}{[Name]}}) \textit{was the Werewolf}}   &  2,679 & 0.962   &  503 & 0.846\\ \hline

\makecell{Mentioned Player \\ Prediction} & \makecell{\textit{I'm the troublemaker and I switched} \\ \textbf{\textcolor{red2}{[Name]}} \textit{with somebody}}   & 3,360 &  N/A & 472 &  N/A \\

\bottomrule
\end{tabular}}
\vspace{-0.1cm}
\captionof{table}{Summary of annotation details for three new social tasks. We achieve sufficiently high \textit{Krippendorff's alpha} values ($\alpha>0.8$) for both speaking target identification and pronoun coreference resolution tasks, which indicates the high reliability of our data annotations.}
\label{table_annotation}}
\vspace{0.0cm}
\end{table*}

\section{Related Work}
\subsection{Social Behavior Analysis}
Analyzing social behaviors has been widely investigated in the fields of computer vision and natural language processing. Various works have focused primarily on analyzing social behaviors from a single-modal perspective. In terms of visual cues, some works proposed gaze target estimation techniques \cite{chong2020detecting, fang2021dual,lai2022eye,tu2022end,tonini2023object} to analyze where a person is looking within a scene. There have also been studies that recognize social gaze patterns between multiple people such as identifying shared attention \cite{hoffman2006probabilistic, fan2018inferring, sumer2020attention, nakatani2023interaction}. Gesture recognition approaches \cite{zhang2018attention,liu2020decoupled,zhou2022decoupling,li2023learning,aich2023data} have been researched to identify specific types of human gestures such as shaking hands and thumbs-up. Regarding language cues, dialogue act recognition methods \cite{tetreault2019dialogue, wang2019persuasion,qin2020dcr,qin2021co,chawla2021casino,malhotra2022speaker} have been introduced to understand the communicative intent behind utterances in social dialogues. Furthermore, there have been works on sentiment analysis and emotion recognition based on dialogue language \cite{jiao2019higru,asai2020emotional,shen2021directed,zhu2021topic,zhao2023knowledge}.

Recently, joint modeling of visual and language modalities has been studied for social behavior analysis. Some works focused on learning the relationships between spoken language and gestures for gesture generation \cite{ahuja2020no, liu2022learning, ahuja2023continual} and gesture-language grounding \cite{lee2021crossmodal}. Liu \textit{et al.} \cite{liu2022learning} proposed a multimodal model that integrates visual, language, and speech cues via hierarchical manners to synthesize naturalistic gestures. Additionally, the intersection of spoken utterances and visual expressions has been explored for sentiment analysis and emotion recognition \cite{zadeh2018multimodal, saha2020towards, hu2022unimse, paraskevopoulos2022mmlatch, hu2022mm, chen2023multivariate}. Hu \textit{et al.} \cite{hu2022unimse} proposed a unified feature space to capture the knowledge of sentiment and emotion comprehensively from multimodal cues. There have also been multimodal works for question \& answering in social contexts \cite{zadeh2019social, xie2023multi, natu2023external}.

However, these works mainly focus on the behaviors of a single person or rely on holistic visual features that are not densely aligned to language in multi-party environments. They are unable to model the complex dynamics of interactions, which requires understanding the spatial relationships of multiple people in addition to their utterances. We propose a novel approach leveraging densely aligned language-visual representations to capture the fine-grained dynamics.

\subsection{Social Deduction Game Modeling}
There have been works investigating computational models for social deduction games where players actively communicate and strategize with one another. Some prior studies have focused on developing game-playing agents and analyzing optimal strategies using game theory \cite{braverman2008mafia, nakamura2016constructing, bi2016human, serrino2019finding, chuchro2022training}. These works aim to model the state of the game computationally but do not address understanding the dialogue and behaviors of players. Chittaranjan \textit{et al.} \cite{chittaranjan2010you} modeled game outcomes from communication patterns such as player speaking and interrupting behaviors. Bakhtin \textit{et al.} \cite{meta2022human} built an agent that can play diplomacy games by utilizing language models with strategic reasoning. These approaches do not capture verbal and non-verbal multimodal aspects of modeling social behaviors. Recently, Lai \textit{et al.} \cite{lai2023werewolf} addressed social behaviors in social deduction games using multimodal representations. They leveraged language and visual cues to predict persuasion strategies at the utterance level such as identity declaration and interrogation.

However, this multimodal work is limited in addressing multi-party dynamics due to the lack of person-level feature recognition. To address this gap, we introduce new tasks in social deduction games that explicitly demand recognizing person-level features. We also propose the corresponding baseline model that captures the multi-party dynamics across both language and visual representations.

\begin{figure*}[t!]
\begin{minipage}[b]{1.0\linewidth}
\centering
\centerline{\includegraphics[width=17.5cm]{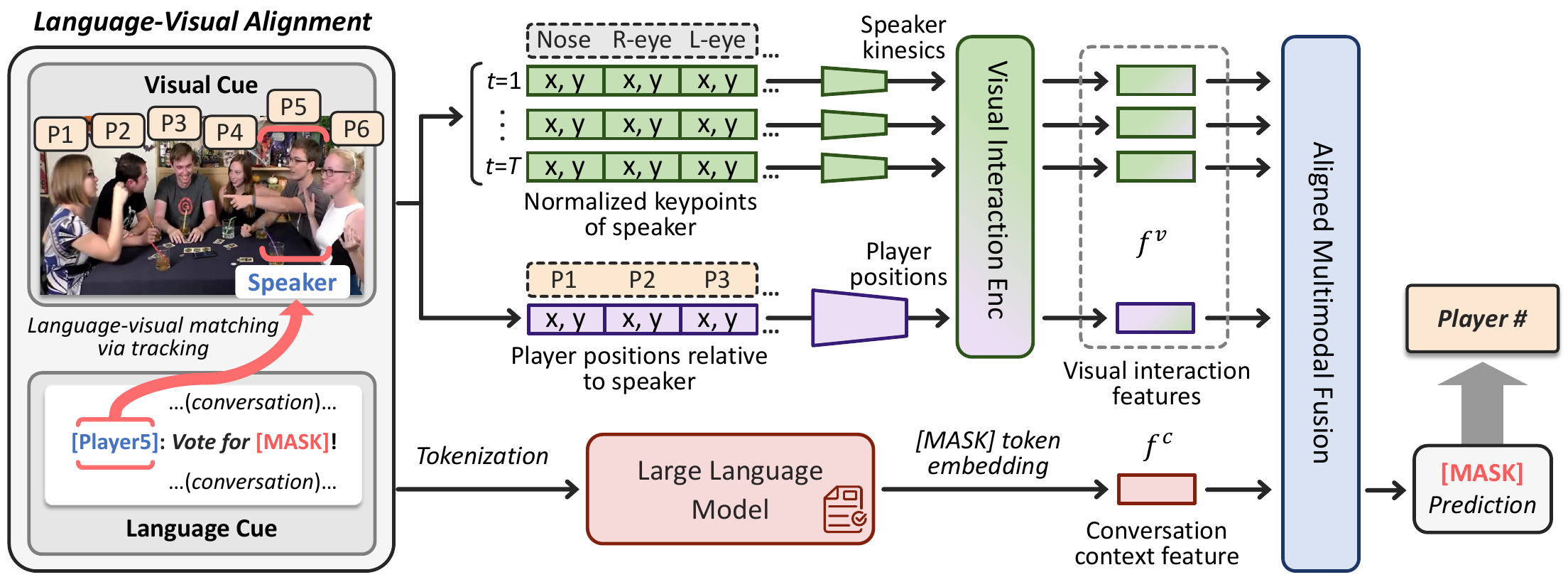}}
\end{minipage}
\vspace{-0.7cm}
\caption{Proposed baseline model for understanding multimodal social interactions to tackle our new social tasks via densely aligned language-visual representations. The model consists of four main parts: language-visual alignment (\textit{grey}), visual interaction modeling (\textit{green} \& \textit{purple}), conversation context modeling (\textit{red}), and aligned multimodal fusion for prediction (\textit{blue}).}
\label{figure_method}
\vspace{-0.30cm}
\end{figure*}

\section{Proposed Benchmark}

\subsection{Base Datasets}
We extend two social deduction game datasets \cite{lai2023werewolf}: YouTube and Ego4D with additional extensive data annotations for curating new social tasks. 

\noindent \textbf{YouTube dataset.}\ This dataset was collected from the YouTube video platform by searching keywords of Werewolf social deduction game. It contains 151 games of One Night Ultimate Werewolf, which corresponds to 151 separate videos with 14.8 hours. It consists of videos, transcripts, player roles, voting outcomes, and persuasion strategy labels. The transcripts comprise 20,832 utterances.

\noindent \textbf{Ego4D dataset.}\ This Ego4D dataset is a subset of Ego4D Social dataset \cite{grauman2022ego4d}. It has 40 games of One Night Ultimate Werewolf and 8 games of The Resistance: Avalon. It contains 101 separate videos with 7.3 hours. Among them, we leverage 83 videos where we can visually identify individuals for new data annotations. To guarantee the visibility of all players within the frame, this dataset adopts third-person view videos instead of first-person view videos. It also consists of videos, transcripts, player roles, voting outcomes, and persuasion strategy labels. The transcripts contain 5,815 utterances during the games.

\subsection{Data Annotation}
To address the fine-grained dynamics of interactions, we design three new tasks in social deduction games: speaking target identification, pronoun coreference resolution, and mentioned player prediction. Annotators reference both transcripts and videos comprehensively to make their annotations in the transcripts. To achieve reliable annotation quality, we initially request three annotators to label subsets of data and measure their annotation agreement using \textit{Krippendorff's alpha} \cite{krippendorff2018content}. After we train the annotators sufficiently with $\alpha$ larger than 0.8, we request the three annotators to label independently for the rest of the data. Note that \textit{Krippendorff's alpha} $>$ 0.8 is generally considered to indicate good reliability with a high level of agreement \cite{carletta22squibs}.

Table \ref{table_annotation} shows the summary of our data annotation results. We achieve sufficiently high $\alpha$ for both speaking target identification and pronoun coreference resolution. Note that we could obtain the annotations for the mentioned player prediction automatically. In the training and testing process, we anonymize all names in transcripts (\textit{e.g.}, [\textit{``David'', ``Alice'', ``Thomas''}] $\rightarrow$ [\textit{``Player1'', ``Player2'', ``Player3''}]). For each task, a test set is constructed using about 20\% of the annotations. We split the training and test sets at the video level rather than at the individual utterance level to ensure no overlap in terms of contextual utterances. Annotation details for each task are as follows.

\noindent \textbf{Task1: Speaking target identification.}\ This task aims to predict who a speaker is talking to in a dialogue. To this end, we annotate the labels of who a speaker is speaking to at the utterance level. Since utterances are often directed to all players, we perform the annotation only on the utterances that include [\textit{``you", ``your"}]. We give our annotators three labeling options: (To Player\#), (To Everyone), and (To Unknown). Based on the annotations, we add ``(To [MASK])" at the end of a target utterance.

\noindent \textbf{Task2: Pronoun coreference resolution.}\ This task aims to predict who a pronoun refers to in a dialogue. We conduct the annotation on the third-person pronouns that are used in our dataset, which are [\textit{``he", ``she", ``him", ``her", ``his"}] in utterances. We give two options to annotators: [Player\#] and [Unknown]. We only target the pronouns that indicate a specific player in the game. In terms of modeling the task, we change a pronoun to [MASK] in a target utterance.

\noindent \textbf{Task3: Mentioned player prediction.}\ This task aims to predict who is referred to by their name in a dialogue. Since we know the ground truth names in utterances, it is possible to annotate these mentioned player labels automatically. We modify a mentioned player name with [MASK] in utterances and predict who is referred to in that part.

\vspace{-0.05cm}
\section{Proposed Approach}

The introduced social tasks can be formulated as follows.  Let $u$$=$$\left\{{u}_{i}\right\}_{i=k-n}^{k+n}$ denote utterance texts that include the $k$-th target utterance containing a [MASK] token representing an unknown player, while $v$$=$$\left\{{v}_{t}\right\}_{t=1}^{T}$ indicates the corresponding $T$ video frames aligned with the utterance timeline. Given $u$ and $v$, our objective is to optimize the multimodal predictive function $\mathcal{F}(v, u)$ to effectively classify the identity of the player associated with the [MASK].

To this end, we introduce a novel multimodal baseline that leverages densely aligned representations between language and visual cues to capture the fine-grained dynamics of interactions. Figure \ref{figure_method} shows the overall framework of the proposed multimodal baseline. The proposed model consists of four main parts: (i) language-visual alignment, (ii) visual interaction modeling, (iii) conversation context modeling, and (iv) aligned multimodal fusion for prediction. 

\subsection{Language-Visual Alignment}
To achieve densely aligned representations containing both verbal and non-verbal dynamics properly, it is necessary to first establish fine-grained alignments between visual and language cues. To this end, we distinguish individual visuals and match language references with them.

We detect and track players visually in video frames over time using AlphaPose framework \cite{alphapose}. Once we initially match player visuals with the player references in the utterances (\textit{i.e.}, assigning each tracking ID to Player\#), we can continuously identify players in both visual and language domains. It enables the model to identify the speaker in the video based on the corresponding utterance and to determine the spatial arrangement of the listeners. Building upon this alignment, we can achieve densely aligned language-visual representations from verbal and non-verbal cues in a comprehensive manner. It enables us to tackle our social tasks effectively, allowing for a more nuanced and holistic understanding of multi-party interactions.

\subsection{Visual Interaction Modeling}
To distinguish individual players in video frames $v$$=$$\left\{{v}_{t}\right\}_{t=1}^{T}$, we use the human pose keypoints from AlphaPose. Specifically, we extract 17 body keypoints ($x$, $y$) for each player. Figure \ref{figure_method} shows the procedure of encoding visual interactions. The upper path (\textit{green}) of Figure \ref{figure_method} indicates encoding a kinesics feature of a speaker, while the middle path (\textit{purple}) represents encoding spatial positions of all players.

First, we use the keypoints of a speaker in the upper path of Figure \ref{figure_method}. Among the 17 part keypoints, we leverage [\textit{nose, l-eye, r-eye, l-shoulder, r-shoulder, l-elbow, r-elbow, l-wrist, r-wrist}] closely related to gaze and gesture characteristics. Let $(x_{part}$, $y_{part})^{S}_{t}$$\in$$\mathbb{R}^{2}$  denote the image coordinates of a part at time $t$. For example, $(x_{nose}$, $y_{nose})^{S}_{t}$ indicates the nose point. To represent human motion in a unified coordinate, we normalize speaker keypoints by subtracting the speaker nose point from each part point. Each point vector $(x_{part}$, $y_{part})^{S}_{t}$ is encoded by an MLP point encoder $E_{point}$ into a part point feature $f^{S, part}_t$$\in$$\mathbb{R}^{d_{point}}$ ($d_{point}$ is channel dim). These part point features are concatenated and processed by an MLP kinesics encoder $E_{kin}$ to obtain a speaker kinesics feature $f^{S}_t$ as follows. 
\begin{equation}
f^{S}_t=E_{kin}([f^{S, nose}_t; f^{S, l\mbox{-}eye}_t; ...; f^{S, r\mbox{-}wrist}_t]). 
\end{equation}
\noindent Since we have multiple time steps, we can obtain $f^{S}$$=$$\left\{{f}_{t}^{S}\right\}_{t=1}^{T}$$\in$$\mathbb{R}^{T \times d}$.

In the meantime, the middle path (\textit{purple}) of Figure \ref{figure_method} receives the position of each player. We consider the nose point of each player as their representative position. We normalize their nose points by subtracting the speaker's nose point from them to get their relative positions from the speaker. We utilize their representative positions at a single time step corresponding to the start of the utterance. Let $(x$, $y)^{P\#}$ denote the representative position of Player\#. Each point vector $(x$, $y)^{P\#}$ is independently fed to an MLP point encoder $E_{point}$ to get a player point feature $f^{P\#}$$\in$$\mathbb{R}^{d_{point}}$. We concatenate the player point features and feed them to an MLP position encoder $E_{pos}$ to get $\tilde{f}^{P}$$\in$$\mathbb{R}^{d}$. We then make $\tilde{f}^{P}$ aware of speaker knowledge. To this end, we make a speaker-label feature $f^{S}_{label}$$\in$$\mathbb{R}^{d}$ by passing a speaker-label one-hot vector through an FC layer. We combine $f^{S}_{label}$ with $\tilde{f}^{P}$ to obtain a player position feature $f^{P}$. These procedures are formulated as follows.
\begin{equation}
\tilde{f}^{P}=E_{pos}([f^{P1}; f^{P2}; ...; f^{PN}]),
\end{equation}
\begin{equation}
f^{P}=\text{FC}(\tilde{f}^{P} + f^{S}_{label}), 
\end{equation}
\noindent where $N$ indicates the maximum player number in the datasets ($N$$=$$6$). If the number of players is less than $N$ for the current input data, we apply zero padding to the excess. If a player is temporarily undetected (\textit{e.g.}, offscreen for a short time), we proceed with position encoding by substituting the corresponding player position stored in a buffer to correct the player position. 

Based on the speaker kinesics features  $f^{S}$$=$$\left\{{f}_{t}^{S}\right\}_{t=1}^{T}$ and player position feature $f^{P}$, we encode the visual interaction by capturing speaker kinesics motion with the context of player visual positions. $f^{S}$ and $f^{P}$ are passed through a visual interaction encoder $E_{v}$ sequentially, which has the form of the transformer \cite{vaswani2017attention}. $E_{{v}}$ allows modeling dependencies between the speaker kinesics and player positions across time via self-attention. Finally, we can obtain visual interaction features $f^{v}$$=$$\left\{{f}_{t}^{v}\right\}_{t=1}^{T+1}$$\in$$\mathbb{R}^{(T+1) \times d}$ that represent dynamics between the speaker and players based on the speaker kinesics and listener positions. 

\subsection{Conversation Context Modeling}
The lower path (\textit{red}) of Figure \ref{figure_method} shows encoding spoken utterances from players. To incorporate conversation context, we use surrounding utterances including the target utterance. The input to the language path is formulated as.
\begin{equation}
u = [u_{k-n}; ...;u_{k-1}; u_{k}; u_{k+1}; ...;u_{k+n}],
\end{equation}
\noindent where $u_k$ denotes the target $k$-th utterance, and the others indicate the preceding and following utterances. Note that the target utterance is the one that contains [MASK]. A [CLS] token is inserted in front of $u$ while a [SEP] token is inserted at the end of each utterance in $u$ for language tokenization processing. Note that all player names in utterances are anonymized as ``[Player\#]". We leverage pre-trained language models based on masked-language modeling such as BERT \cite{kenton2019bert}. The tokenized sequence of the utterances is fed into the language model. The output feature corresponding to the index of the [MASK] token is then retrieved. After passing it through an FC layer to match the channel dimension of the visual interaction features, we get a conversation context feature $f^{c}$$\in$$\mathbb{R}^{d}$ which contains the context around the [MASK].

\begin{table}[t]{
\renewcommand{\arraystretch}{1.2}
\renewcommand{\tabcolsep}{0.6mm}
\centering
\resizebox{0.999\linewidth}{!}{
\begin{tabular}{l  c  P{1.4cm} P{1.4cm}}
\toprule
\multirow{2}{*}{\bf Method} & \multirow{2}{*}{\bf \makecell{Densely \\Aligned?}} & \multicolumn{2}{c}{\bf \makecell{Speaking Target \\ Identification (\%)}}  \\ 
\cmidrule(lr){3-4}
& & \bf \makecell{YouTube} & \bf \makecell{Ego4D}  \\ \bottomrule

\makecell[l]{BERT \cite{kenton2019bert}}  & -   & 65.8 &  56.8 \\ 
\makecell[l]{BERT + DINOv2 \cite{oquab2023dinov2}} & \xmark   & 66.4 &  58.0 \\ 
\makecell[l]{BERT + MViT \cite{fan2021multiscale} (Lai \textit{et al.} \cite{lai2023werewolf})} & \xmark   & 66.9 &  57.4  \\ 
{\bf \graycell BERT-based Our Baseline} & \graycell \cmark   & \bf \graycell 72.7 &  \bf \graycell 61.9  \\ \hline 

\makecell[l]{RoBERTa \cite{liu2019roberta}} & -   & 72.4 &  63.6  \\ 
\makecell[l]{RoBERTa + DINOv2 \cite{oquab2023dinov2}} & \xmark   & 72.7 &  62.5 \\ 
\makecell[l]{RoBERTa + MViT \cite{fan2021multiscale} (Lai \textit{et al.} \cite{lai2023werewolf})} & \xmark   & 73.1 &  64.2  \\ 
{\bf \graycell RoBERTa-based Our Baseline} & \graycell \cmark   & \bf \graycell 74.5 &  \bf \graycell66.5  \\ \hline 

\makecell[l]{ELECTRA \cite{clark2020electra}} & -   & 65.8 &  60.8  \\ 
\makecell[l]{ELECTRA + DINOv2 \cite{oquab2023dinov2}} & \xmark   & 65.3 &  60.2 \\ 
\makecell[l]{ELECTRA + MViT \cite{fan2021multiscale} (Lai \textit{et al.} \cite{lai2023werewolf})} & \xmark   & 64.6 &  60.8  \\ 
{\bf \graycell ELECTRA-based Our Baseline} & \graycell \cmark   & \bf \graycell 69.6 &  \bf \graycell 64.8  \\ \toprule

\end{tabular}}
\vspace{-0.1cm}
\captionof{table}{Performance comparison results for the speaking target identification task on YouTube and Ego4D datasets.}
\label{table_performance1}}
\vspace{-0.1cm}
\end{table}

\subsection{Aligned Multimodal Fusion}

To fuse the aligned visual interaction features $f^{v}$$=$$\left\{{f}_{t}^{v}\right\}_{t=1}^{T+1}$ and conversation context feature $f^{c}$, we first concatenate them in the sequence dimension along with an [AGG] token for feature aggregation. It can be formulated as follows.
\begin{equation}
f^{v+c} = [\text{[AGG]}; f_1^{v}; ...;f_{T+1}^{v}; f^{c}].
\end{equation}
Note that positional encoding \cite{vaswani2017attention} for transformers is applied to the $f^{v}$ parts. Then, $f^{v+c}$$\in$$\mathbb{R}^{(T+2) \times d}$ is processed with a multimodal transformer to encode their joint relationships. We leverage an output multimodal feature $f^{m}$$\in$$\mathbb{R}^{d}$ from the [AGG] token. Finally, a densely aligned multimodal feature $f^{m}$ is passed through a classification head consisting of an FC layer and softmax to predict the anonymized player identity $\hat{y}$ (\textit{e.g.}, Player\#) for the target [MASK]. We optimize the model using cross-entropy loss between the predicted player $\hat{y}$ and the ground-truth label $y$. 

At training time, we apply permutations to anonymized identities to prevent the model from relying on consistent identities. Specifically, we randomly shuffle the mapping from player names to the anonymized player identities in utterances for every iteration. For example, [\textit{``David'', ``Alice'', ``Thomas''}] $\rightarrow$ [\textit{``Player1'', ``Player2'', ``Player3''}] $\rightarrow$ [\textit{``Player3'', ``Player1'', ``Player2''}]. This mapping permutation from the text domain is also applied to the visual position encoding and ground truth label $y$ to ensure that language and visual cues are consistently aligned. This player permutation learning forces the model to learn more generalizable representations of player interactions that do not depend on specific identifiers during the training time.

\begin{table}[t]{
\renewcommand{\arraystretch}{1.2}
\renewcommand{\tabcolsep}{0.1mm}
\centering
\resizebox{0.999\linewidth}{!}{
\begin{tabular}{l  c  P{1.6cm} P{1.6cm}}
\toprule
\multirow{2}{*}{\bf Method} & \multirow{2}{*}{\bf \makecell{Densely \\Aligned?}} & \multicolumn{2}{c}{\bf \makecell{Pronoun Coreference\\ Resolution (\%)}}  \\ 
\cmidrule(lr){3-4}
& & \bf {YouTube} & \bf {Ego4D}   \\ \bottomrule

\makecell[l]{BERT \cite{kenton2019bert}} & -   & 60.3 &  47.3 \\ 
\makecell[l]{BERT + DINOv2 \cite{oquab2023dinov2}} & \xmark   & 58.2 &  46.4 \\ 
\makecell[l]{BERT + MViT \cite{fan2021multiscale} (Lai \textit{et al.} \cite{lai2023werewolf})} & \xmark   & 59.8 &  46.4  \\ 
{\bf \graycell BERT-based Our Baseline} & \graycell \cmark   & \bf \graycell 65.9 &  \bf\graycell49.1  \\ \hline 

\makecell[l]{RoBERTa \cite{liu2019roberta}} & -   & 69.0 &  48.2  \\ 
\makecell[l]{RoBERTa + DINOv2 \cite{oquab2023dinov2}} & \xmark   & 68.6 &  46.4 \\ 
\makecell[l]{RoBERTa + MViT \cite{fan2021multiscale} (Lai \textit{et al.} \cite{lai2023werewolf})} & \xmark   & 69.5 &  49.1  \\ 
{\bf \graycell RoBERTa-based Our Baseline} & \graycell \cmark   & \bf \graycell73.0 &  \bf \graycell52.7  \\ \hline 

\makecell[l]{ELECTRA \cite{clark2020electra}} & -   & 62.5 &  44.6  \\ 
\makecell[l]{ELECTRA + DINOv2 \cite{oquab2023dinov2}} & \xmark   & 61.1 &  42.9 \\ 
\makecell[l]{ELECTRA + MViT \cite{fan2021multiscale} (Lai \textit{et al.} \cite{lai2023werewolf})} & \xmark   & 62.1 &  43.8  \\ 
{\bf \graycell ELECTRA-based Our Baseline} & \graycell \cmark   & \bf \graycell67.6 & \bf \graycell 46.4  \\ 

\toprule
\end{tabular}}
\vspace{-0.1cm}
\captionof{table}{Performance comparison results for the pronoun coreference resolution task on YouTube and Ego4D datasets.}
\label{table_performance2}}
\vspace{-0.1cm}
\end{table}

\section{Experiments}
\subsection{Implementation}
We adopt the language model as pre-trained BERT \cite{kenton2019bert}, RoBERTa \cite{liu2019roberta}, and ELECTRA \cite{clark2020electra} which are based on masked-language modeling. The proposed model is trained by Adam optimizer \cite{kingma2015adam} with a learning rate of 5e-6 for the language model and 5e-5 for the other parts. We use a batch size of 16. We leverage about 3 seconds of video frames (frame interval 0.4s) that correspond to the timeline of the utterance. We use the preceding and following 5 utterances for encoding conversation context. The detailed network structures are described in the supplementary material.

\subsection{Performance Comparison}
We measure the identity classification accuracies for our curated tasks: speaking target identification, pronoun coreference resolution, and mentioned player prediction. 

Table \ref{table_performance1} shows the experimental results for speaking target identification on YouTube and Ego4D datasets with different language models. We compare our proposed baselines with the recent multimodal model \cite{lai2023werewolf} (\textit{i.e.,} Language Model + MViT \cite{fan2021multiscale}) for social deduction games. In addition, we further adopt DINOv2 \cite{oquab2023dinov2} which is a powerful versatile visual feature generally used for various downstream tasks. Note that both comparison methods cannot leverage densely aligned language-visual representations. As shown in the table, these methods are not effective in improving upon the language models alone. This reflects that they are not able to figure out who the speaker is and who their gestures are directed at, in correspondence with the language domain. In contrast, our baselines leveraging densely aligned language-visual representations consistently enhance the language models for this task. 

Table \ref{table_performance2} and \ref{table_performance3} show the performance comparison results for pronoun coreference resolution and mentioned player prediction, respectively. We follow a similar experimental setup, evaluating our baselines against the language models and the multimodal methods (\textit{i.e.}, Language Models + DINOv2/MViT) across three different language models and two datasets. The results show that the competing multimodal methods fail to achieve substantial improvements over the language baselines. In contrast, our proposed multimodal baseline consistently outperforms both the language models and the other multimodal methods. Our multimodal approach demonstrates the effectiveness of aligned multimodal cues in addressing these social tasks.

\begin{table}[t]{
\renewcommand{\arraystretch}{1.15}
\renewcommand{\tabcolsep}{0.2mm}
\centering
\resizebox{0.999\linewidth}{!}{
\begin{tabular}{l  c  P{1.4cm} P{1.4cm}}
\toprule
\multirow{2}{*}{\bf Method} & \multirow{2}{*}{\bf \makecell{Densely \\Aligned?}} & \multicolumn{2}{c}{\bf \makecell{Mentioned Player \\ Prediction (\%)}}  \\ 
\cmidrule(lr){3-4}
& & \bf \makecell{YouTube} & \bf \makecell{Ego4D}  \\ \bottomrule

\makecell[l]{BERT \cite{kenton2019bert}} & -   & 54.6 &  46.2 \\ 
\makecell[l]{BERT + DINOv2 \cite{oquab2023dinov2}} & \xmark   & 54.4 &  47.4 \\ 
\makecell[l]{BERT + MViT \cite{fan2021multiscale} (Lai \textit{et al.} \cite{lai2023werewolf})} & \xmark   & 53.1 &  46.2  \\ 
{\bf \graycell BERT-based Our Baseline} & \graycell \cmark   & \bf \graycell 58.8 & \bf\graycell 50.0  \\ \hline 

\makecell[l]{RoBERTa \cite{liu2019roberta}} & -   & 59.9 &  50.0  \\ 
\makecell[l]{RoBERTa + DINOv2 \cite{oquab2023dinov2}} & \xmark   & 60.7 &  50.0 \\ 
\makecell[l]{RoBERTa + MViT \cite{fan2021multiscale} (Lai \textit{et al.} \cite{lai2023werewolf})} & \xmark   & 60.6 &  51.3  \\ 
{\bf \graycell RoBERTa-based Our Baseline} & \graycell \cmark   &\bf \graycell62.5 & \bf \graycell55.1  \\ \hline 

\makecell[l]{ELECTRA \cite{clark2020electra}} & -   & 55.7 &  42.3  \\ 
\makecell[l]{ELECTRA + DINOv2 \cite{oquab2023dinov2}} & \xmark   & 56.1 &  43.6 \\ 
\makecell[l]{ELECTRA + MViT \cite{fan2021multiscale} (Lai \textit{et al.} \cite{lai2023werewolf})} & \xmark   & 55.3 &  41.0  \\ 
{\bf \graycell ELECTRA-based Our Baseline} & \graycell \cmark   &\bf \graycell61.0 &  \bf\graycell46.2  \\ 

\toprule
\end{tabular}}
\vspace{-0.1cm}
\captionof{table}{Performance comparison results for the mentioned player prediction task on YouTube and Ego4D datasets.}
\label{table_performance3}}
\vspace{-0.3cm}
\end{table}

\subsection{Effects of Visual Features}
We conduct ablation studies on visual feature types to analyze the contribution of each component in our baseline model. Table \ref{table_ablation1} shows the performance results according to the types of encoded non-verbal cues (\textit{i.e.}, gesture and gaze features) for our social tasks. Our final baseline model encodes speaker kinesics using keypoints related to gaze and gesture, specifically [\textit{nose, l-eye, r-eye, l-shoulder, r-shoulder, l-elbow, r-elbow, l-wrist, r-wrist}]. ``w/o gesture feature" indicates the model utilizing only the head-related keypoints of [\textit{nose, l-eye, r-eye}] while  ``w/o gaze feature" employs only the gesture-related keypoints of [\textit{l-shoulder, r-shoulder, l-elbow, r-elbow, l-wrist, r-wrist}]. We adopt the BERT-based baseline for evaluation on YouTube dataset. As shown in the table, the results show that the gesture features are more dominant compared to the gaze features in our setting. The proposed baselines using both gesture and gaze features generally achieve good performances. 

\begin{table}[t]{
\renewcommand{\arraystretch}{1.15}
\renewcommand{\tabcolsep}{3.0mm}
\centering
\resizebox{0.999\linewidth}{!}{
\begin{tabular}{c  l c  c }
\toprule
\multirow{1}{*}{\bf Target Task} & {\bf {Method}} & \bf \multirow{1}{*}{Accuracy (\%)}  \\ \bottomrule 

\multirow{4}{*}{\makecell{ Speaking Target \\ Identification}} & \makecell[l]{ w/o Visual Features }   & 65.8   \\
& \makecell[l]{ w/o Gesture Feature }   & 69.6    \\ 
& \makecell[l]{ w/o Gaze Feature }   & 70.2   \\   
&{\bf\graycell Our Baseline }   & \bf \graycell 72.7   \\\hline

\multirow{4}{*}{\makecell{ Pronoun Coreference \\ Resolution}}  & \makecell[l]  {w/o Visual Features}   & 60.3   \\
& \makecell[l]{ w/o Gesture Feature }    & 64.9   \\  
& \makecell[l]{ w/o Gaze Feature }    &\bf 66.5   \\ 
&{\bf\graycell Our Baseline }    & \underline{\graycell65.9}   \\ \hline

\multirow{4}{*}{\makecell{ Mentioned Player \\ Prediction}}  & \makecell[l] {w/o Visual Features}    & 54.6   \\
&\makecell[l]{w/o Gesture Feature } &55.8  \\  
&{w/o Gaze Feature } & 56.2   \\  
&{\bf\graycell Our Baseline } & \bf \graycell 58.8   \\ 

\toprule
\end{tabular}}
\vspace{-0.1cm}
\captionof{table}{Effects of non-verbal visual feature types on the performances for three social tasks. We adopt the BERT-based model and conduct evaluations on YouTube dataset.}
\label{table_ablation1}}
\end{table}

\begin{table}[t]{
\renewcommand{\arraystretch}{1.15}
\renewcommand{\tabcolsep}{2mm}
\centering
\resizebox{0.999\linewidth}{!}{
\begin{tabular}{c  c c  c }
\toprule
\multirow{2}{*}{\bf Target Task} & \multicolumn{2}{c}{\bf \makecell{Conversation Context}} &  \bf \multirow{2}{*}{Accuracy (\%)}  \\ 
\cmidrule(lr){2-3}
&  {\bf \makecell{Preceding}} & {\bf \makecell{Following }}  & \\ \bottomrule 

\multirow{3}{*}{\makecell{ Speaking Target \\ Identification}} &  \xmark & \xmark  &40.2   \\
& \cmark & \xmark   &  59.1   \\ 
& \graycell \cmark & \graycell \cmark   &  \bf \graycell72.7   \\  \hline

\multirow{3}{*}{\makecell{ Pronoun Coreference \\ Resolution}} & \xmark & \xmark  & 51.1   \\
& \cmark & \xmark   &  63.4   \\  
&  \graycell \cmark & \graycell \cmark   &   \bf \graycell65.9   \\  \hline

\multirow{3}{*}{\makecell{ Mentioned Player \\ Prediction}} &  \xmark & \xmark  &  36.2   \\
& \cmark & \xmark    & 47.3   \\  
&  \graycell \cmark & \graycell \cmark    &\bf \graycell 58.8   \\  

\toprule
\end{tabular}}
\vspace{-0.1cm}
\captionof{table}{Effects of the language conversation contexts on the performances for three social tasks. We adopt the BERT-based model and conduct evaluations on YouTube dataset.}
\label{table_context}}
\vspace{-0.3cm}
\end{table}

\begin{figure*}[t!]
\begin{minipage}[b]{1.0\linewidth}
\centering
\centerline{\includegraphics[width=17.5cm]{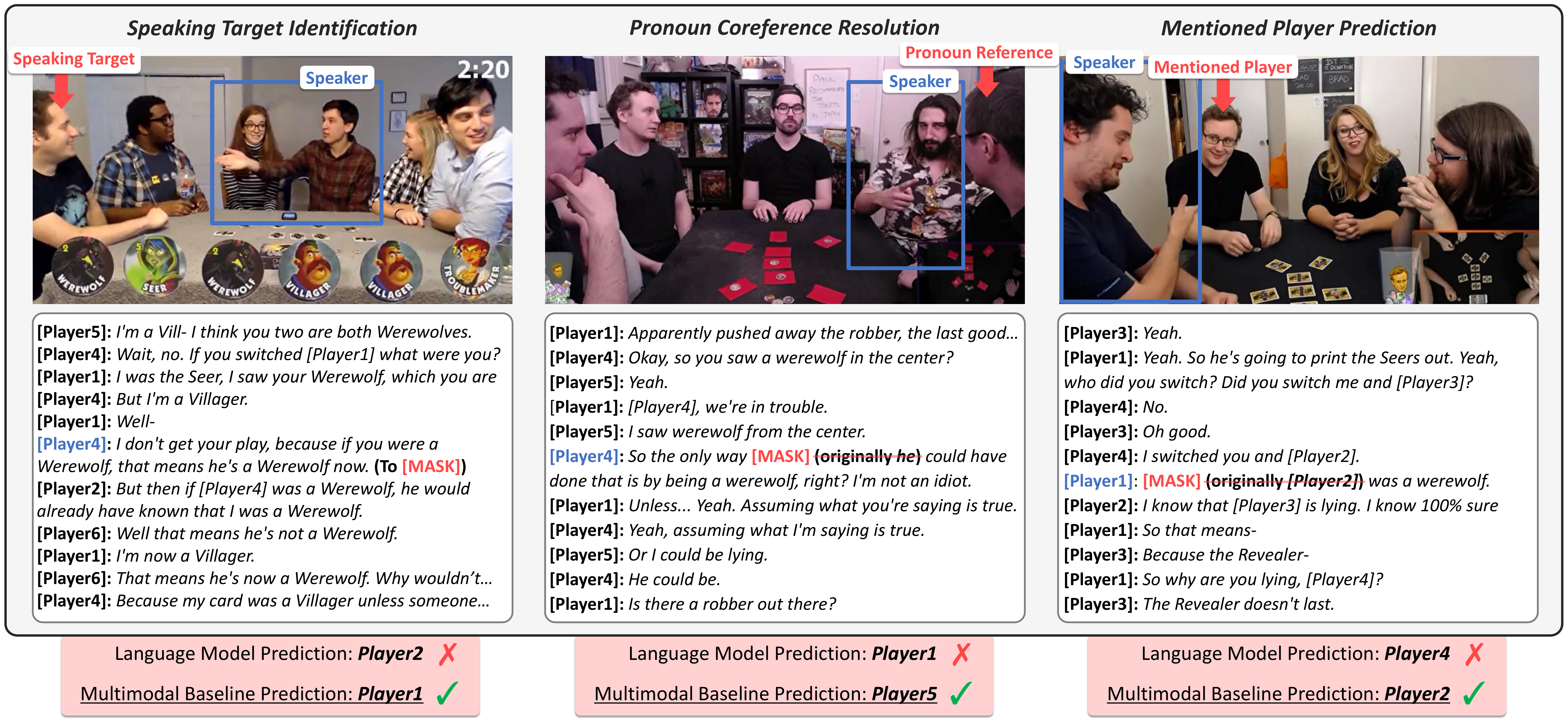}}
\end{minipage}
\vspace{-0.6cm}
\caption{Qualitative results demonstrating the benefit of visual cues for three social tasks. The examples show cases where the language model alone fails, but the proposed multimodal baseline leveraging both language and visual cues correctly predicts the right person. Note that Player\# are assigned in ascending order from left to right in the visual scenes of this figure.}
\label{figure_qualitative}
\vspace{-0.20cm}
\end{figure*}

\subsection{Effects of Conversation Context}
Conversational context plays an important role in understanding the meaning and intent behind individual utterances. To analyze the effects of conversation context on our social tasks, we conduct ablation experiments according to the preceding and following contexts of the target utterance. Table \ref{table_context} presents the results of different context settings using the BERT-based baseline model on YouTube dataset. We compare three variations of the model: one that uses only the target utterance without any additional context, another that incorporates only the preceding context, and our proposed baseline which utilizes both preceding and following contexts. As shown in the table, leveraging both contexts shows the best results for all social tasks. It is noteworthy that the advantage of using the following context is relatively small in the pronoun coreference resolution task compared to the other tasks. This is because the nature of this task is to resolve the reference of pronouns against people that usually appear in the preceding context.

\begin{table}[t]{
\renewcommand{\arraystretch}{1.2}
\renewcommand{\tabcolsep}{2.6mm}
\centering
\resizebox{0.999\linewidth}{!}{
\begin{tabular}{c  c  c }
\toprule
\multirow{1}{*}{\bf Target Task} & {\bf \makecell{Player Permutation \\ Learning}} &  \bf \multirow{1}{*}{Accuracy (\%)}  \\  \bottomrule 
\multirow{2}{*}{\makecell{ Speaking Target \\ Identification}} & \xmark   &  70.8   \\  
& \graycell\cmark   &  \bf \graycell 72.7   \\ \hline
\multirow{2}{*}{\makecell{ Pronoun Coreference \\ Resolution}} &  \xmark   & 51.5  \\  
& \graycell\cmark    &  \bf \graycell65.9    \\  \hline
\multirow{2}{*}{\makecell{ Mentioned Player \\ Prediction}} &  \xmark   & 52.7   \\  
&  \graycell \cmark    & \bf \graycell 58.8   \\  
\toprule
\end{tabular}}
\vspace{-0.1cm}
\captionof{table}{Effects of player permutation learning on the performances for three social tasks. We adopt the BERT-based model and conduct evaluations on YouTube dataset.}
\label{table_permutation}}
\vspace{-0.1cm}
\end{table}

\subsection{Effects of Permutation Learning}
To validate the effectiveness of our player permutation learning which shuffles anonymized player identities, we conduct ablation experiments by training models with and without permutation. Table \ref{table_permutation} shows the experiment results for three tasks with our BERT-based baseline on YouTube dataset. As shown in the table, the permutation learning approach consistently improves the performances for all tasks, implying it helps the model learn more generalizable representations of player interactions. Note that we apply this player permutation learning to all comparison methods in Tables \ref{table_performance1}, \ref{table_performance2}, and \ref{table_performance3} for fair performance comparisons.

\subsection{Qualitative Results}
Figure \ref{figure_qualitative} shows examples of three social tasks and their qualitative results according to the use of visual cues. We utilize BERT as the language model for this experiment. As shown in the figure, our multimodal baseline leveraging both language and visual cues in a dense alignment manner can correct the inference when the language model alone fails. The qualitative results show that visual features aligned to utterances provide complementary information to disambiguate referents in social interactions.

\section{Conclusion}
We introduce three new challenging tasks in social deduction games: speaking target identification, pronoun coreference resolution, and mentioned player prediction - all of which require understanding the fine-grained verbal and non-verbal dynamics between multiple people. We curate extensive dataset annotations for our new social tasks and further propose a novel multimodal baseline that establishes dense language-visual alignments between spoken utterances and player visual features. This approach enables modeling multi-party social interactions through verbal and non-verbal communication channels simultaneously. Experiments show consistent and considerable performance improvements of our multimodal baselines over other approaches without both modalities and without multimodal dense alignment. Furthermore, extensive ablation studies are conducted to validate the effectiveness of our baseline components. We release the benchmarks and source code to facilitate further research in this direction. \\

\vspace{-0.2cm}
\noindent{\bf Acknowledgement.} 
Portions of this project were supported in part by a gift from Meta.

{
    \small
    \bibliographystyle{ieeenat_fullname}
    \bibliography{refs}
}

\clearpage
\appendix
\onecolumn
\vspace*{0.1cm}

\begin{center}
{\fontsize{14.5}{14.5}{\textbf {Modeling Multimodal Social Interactions: New Challenges and Baselines}}} \\ \vspace{0.18cm}
{\fontsize{14.5}{14.5}{\textbf {with Densely Aligned Representations}}}  \\ \vspace{0.18cm}
{\fontsize{14.5}{14.5}{\textbf { - \textit{Supplementary Material} -}}} \\
\vspace{0.9cm} 

\end{center}

\vspace{1.2cm}
\section{Network Structure Details}
Table \ref{table_network} shows the network structure details of the proposed baseline. The point, kinesics, position, and visual interaction encoders are included in the visual interaction modeling part. The multimodal transformer is used for aligned multimodal fusion. The point, kinesics, and position encoders are implemented with Multilayer Perceptron (MLP) structures, comprising fully connected (FC) layers. The visual interaction encoder and the multimodal transformer are based on typical transformer architectures  \cite{vaswani2017attention}. For the MLP structures, ``MLP Size" denotes the output dimension of each FC layer. ReLU activation is applied between FC layers in MLP structures. For transformers, ``Hidden Size” refers to the feature size after passing through the feed-forward network, while ``MLP Size” represents the intermediate feature size in the Multi-Head Attention (MHA) mechanism. The channel dimensions $d_{point}$ and $d$ are set to 64 and 512, respectively.


\begin{table}[h]
	\renewcommand{\arraystretch}{0.6}
	\renewcommand{\tabcolsep}{3mm}
		\vspace{0.5cm}
	\centering
	\begin{center}
		\resizebox{0.75\linewidth}{!}
		{
			\begin{tabular}{M{4cm}M{2cm}M{2cm}M{2cm}M{2cm}}
				\toprule
				
				\multicolumn{5}{c}{\textbf{Network Structures}}  \\
				\cmidrule(lr){1-5}

				\multirow{2}{*}{\textbf{Module}} & \multirow{2}{*}{\textbf{Layers}} & \multirow{2}{*}{\textbf{Hidden Size}} & \multirow{2}{*}{\textbf{MLP Size}} & \multirow{2}{*}{\textbf{Multi-Heads}}  \\
				&  & &  &  \\ \toprule
				
				\multirow{4}{*}{\makecell{Point Encoder }} & \multirow{4}{*}{3} & \multirow{4}{*}{--} & \multirow{4}{*}{64}  &  \multirow{4}{*}{--}\\
				&  &  &  &  \\ 
				&  &  &  &   \\
				&  &  &  &   \\  \hdashline
				
				\multirow{4}{*}{\makecell{Kinesics Encoder }} & \multirow{4}{*}{4} & \multirow{4}{*}{--} & \multirow{4}{*}{512}  &  \multirow{4}{*}{--}\\
				&  &  &  &  \\ 
				&  &  &  &   \\
				&  &  &  &   \\  \hdashline

				\multirow{4}{*}{\makecell{Position Encoder }} & \multirow{4}{*}{4} & \multirow{4}{*}{--} & \multirow{4}{*}{512}  &  \multirow{4}{*}{--}\\
				&  &  &  &  \\ 
				&  &  &  &   \\
				&  &  &  &   \\  \hdashline

				\rule{0pt}{10pt}\multirow{4}{*}{\makecell{Visual Interaction \\ Encoder}} & \multirow{4}{*}{3} & \multirow{4}{*}{512} & \multirow{4}{*}{1024}  &  \multirow{4}{*}{8}\\
				&  &  &  &  \\ 
				&  &  &  &   \\
				&  &  &  &   \\  \hdashline

				\rule{0pt}{10pt}\multirow{4}{*}{\makecell{Aligned Multimodal    \\ Transformer }} & \multirow{4}{*}{2} & \multirow{4}{*}{512} & \multirow{4}{*}{1024}  &  \multirow{4}{*}{8}\\
				&  &  &  &  \\ 
				&  &  &  &   \\
				&  &  &  &   \\ 
    			
				\bottomrule
			\end{tabular}
		}
  \caption{Network structure details of the proposed baseline model including MLP and transformer structures.}
  \label{table_network}
	\end{center}
	\vspace{1.0cm}
	
\end{table}

\newpage
\section{Implementation Details}
\textbf{Language Models.} We employ "bert-base-uncased", ``roberta-base", and ``electra-base-discriminator" as pre-trained BERT, RoBERTa, and ELECTRA language models, respectively. We leverage models and weights of them from Hugging Face \cite{wolf2020transformers}. 

\noindent \textbf{Comparison Methods.} For comparative analysis, we utilize (Language Model + MViT) and (Language Model + DINOv2). In the case of (Language Model + MViT), we leverage the visual features from the 24-layer multiscale vision transformer (MViT) \cite{fan2021multiscale}, pre-trained on the Kinetics-400 video dataset, following the approach in \cite{lai2023werewolf}. For (Language Model + DINOv2), we use the visual features pooled from a 3-second window of DINOv2 features \cite{oquab2023dinov2} (interval 0.5s) along the time axis. Both visual features are integrated with the language feature (\textit{i.e.}, conversation context feature) through FC layers for task-specific predictions according to \cite{lai2023werewolf}.

\vspace{0.5cm}
\section{Effects of Conversation Context Length}
\begin{figure}[h]
    \begin{minipage}[b]{0.49\textwidth}
        \centering
        \includegraphics[width=0.8\linewidth]{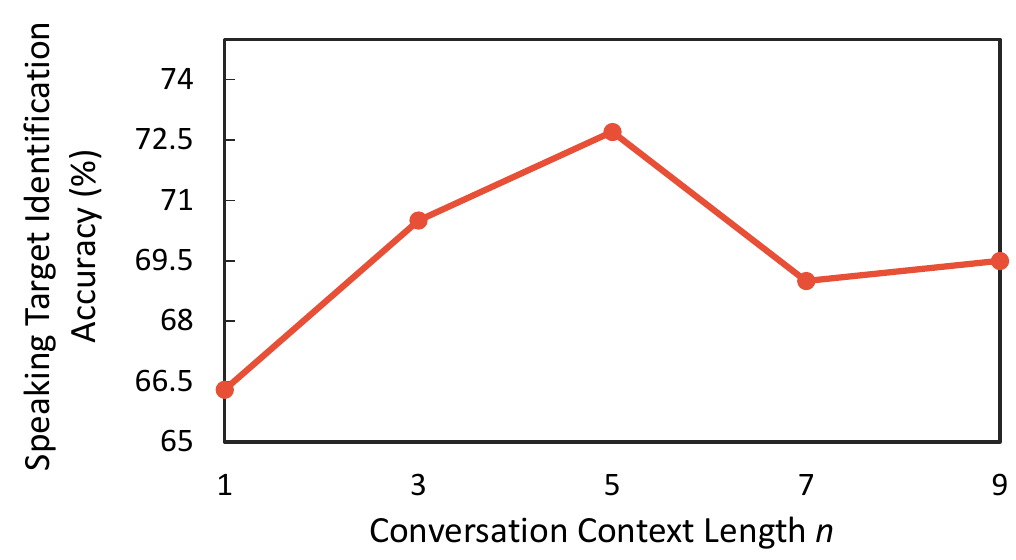}
        \vspace{-0.4cm}
    \end{minipage}
    \begin{minipage}[b]{0.5\textwidth}
    		\renewcommand{\arraystretch}{1.8}
		\renewcommand{\tabcolsep}{5mm}
        \centering
        \resizebox{0.8\linewidth}{!}{%
        \begin{tabular}{c c c}
				\toprule
				 \multirow{1}{*}{\bf Target Task} & {\bf \makecell{Conversation \\  Context Length}} &  \bf \multirow{1}{*}{Accuracy (\%)}  \\  \bottomrule 
      \multirow{5}{*}{\makecell{ Speaking Target \\ Identification}} & $n=1$   &   66.3 \\ 
    & $n=3$   &  70.5  \\  

    & \bm{${n=5}$}  \graycell & \bf \graycell 72.7 \\ 
    & $n=7$    &    69.0  \\  

     &  $n=9$    & 69.5  \\  
				\bottomrule
        \end{tabular}
        }
        \vspace{-0.1cm}
    \end{minipage}%
        \captionof{figure}{Effects of conversation context length on the performance for speaking target identification.}
        \label{figure_context1}
\end{figure}

\begin{figure}[h]
    \begin{minipage}[b]{0.49\textwidth}
        \centering
        \includegraphics[width=0.8\linewidth]{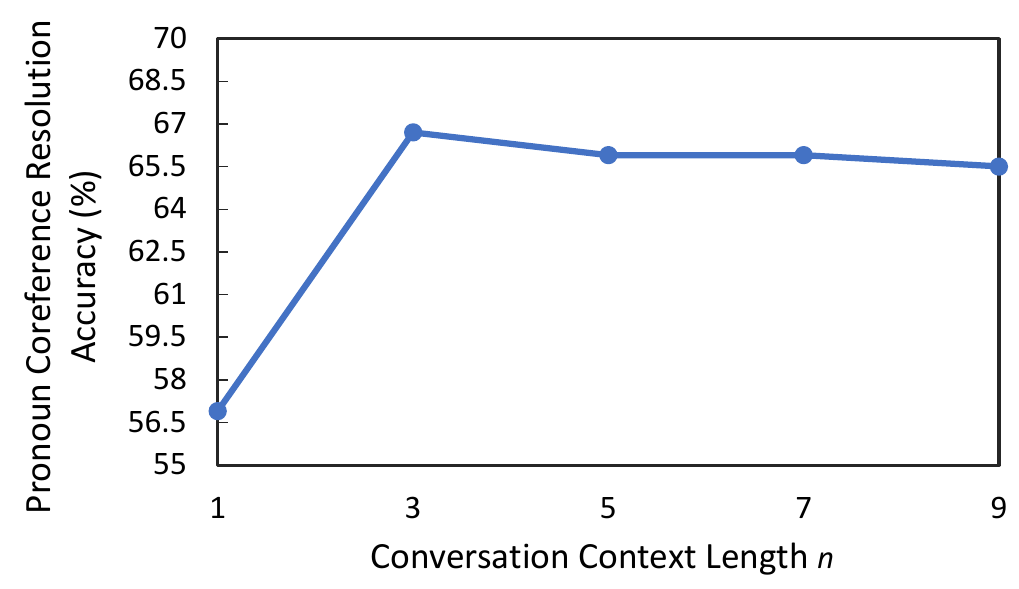}
        \vspace{-0.4cm}
    \end{minipage}
    \begin{minipage}[b]{0.5\textwidth}
    		\renewcommand{\arraystretch}{1.8}
		\renewcommand{\tabcolsep}{5mm}
        \centering
        \resizebox{0.8\linewidth}{!}{%
        \begin{tabular}{c c c}
				\toprule
				 \multirow{1}{*}{\bf Target Task} & {\bf \makecell{Conversation \\  Context Length}} &  \bf \multirow{1}{*}{Accuracy (\%)}  \\  \bottomrule 
      \multirow{5}{*}{\makecell{ Pronoun Coreference \\ Resolution}} & $n=1$   &  56.9  \\ 
    & $n=3$   & \bf 66.7   \\  

    &  \bm{${n=5}$}  \graycell   &  \underline{\graycell65.9}  \\ 
    & $n=7$    &    65.9  \\  

     &  $n=9$    &  65.5  \\  
				\bottomrule
        \end{tabular}
        }
        \vspace{-0.1cm}
    \end{minipage}%
        \captionof{figure}{Effects of conversation context length on the performance for pronoun coreference resolution.}
        \label{figure_context2}
\end{figure}

\begin{figure}[h]
    \begin{minipage}[b]{0.49\textwidth}
        \centering
        \includegraphics[width=0.8\linewidth]{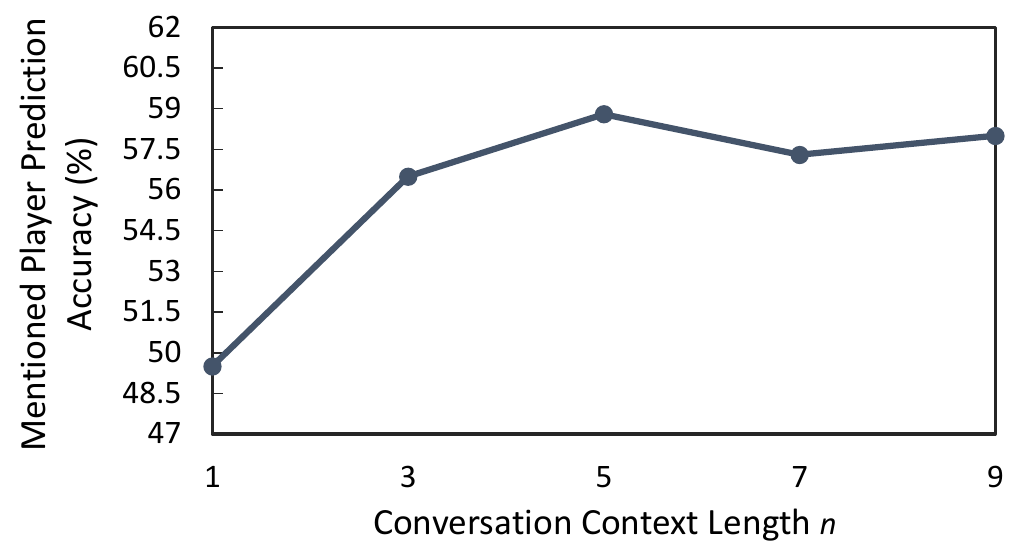}
        \vspace{-0.4cm}
    \end{minipage}
    \begin{minipage}[b]{0.5\textwidth}
    		\renewcommand{\arraystretch}{1.8}
		\renewcommand{\tabcolsep}{5mm}
        \centering
        \resizebox{0.8\linewidth}{!}{%
        \begin{tabular}{c c c}
				\toprule
				 \multirow{1}{*}{\bf Target Task} & {\bf \makecell{Conversation \\ Context Length}} &  \bf \multirow{1}{*}{Accuracy (\%)}  \\  \bottomrule 
      \multirow{5}{*}{\makecell{ Mentioned Player \\ Prediction}} & $n=1$   &  49.5  \\
    & $n=3$   &  56.5   \\  

    &  \bm{${n=5}$}  \graycell   & \bf \graycell 58.8 \\ 
    & $n=7$    &  57.3    \\  

     &  $n=9$    &  58.0 \\  
				\bottomrule
        \end{tabular}
        }
        \vspace{-0.1cm}
    \end{minipage}%
        \captionof{figure}{Effects of conversation context length on the performance for mentioned player prediction.}
        \label{figure_context3}
\end{figure}

We conduct experiments to investigate the effects of conversation context length $n$ on the performance of each task. Figures \ref{figure_context1}, \ref{figure_context2}, and \ref{figure_context3} show the validation results for speaking target identification, pronoun coreference resolution, and mentioned player prediction, respectively. When a context length of $n$ is employed, the target utterance is concatenated with $n$ preceding and $n$ following utterances. As shown in the figures, we consistently obtain low performances with the shortest context length of $n=1$, while achieving fairly good performances with a context length of $n=5$ for all tasks. Note that we adopt $n=5$ as our default setting for the baselines. These evaluations were conducted on YouTube dataset using the BERT-based model.

\vspace{0.5cm}
\vspace{-0.8cm}
\vspace{-0.8cm}

\newpage
\section{Effects of Video Length}
\begin{figure}[h]
    \begin{minipage}[b]{0.49\textwidth}
        \centering
        \includegraphics[width=0.8\linewidth]{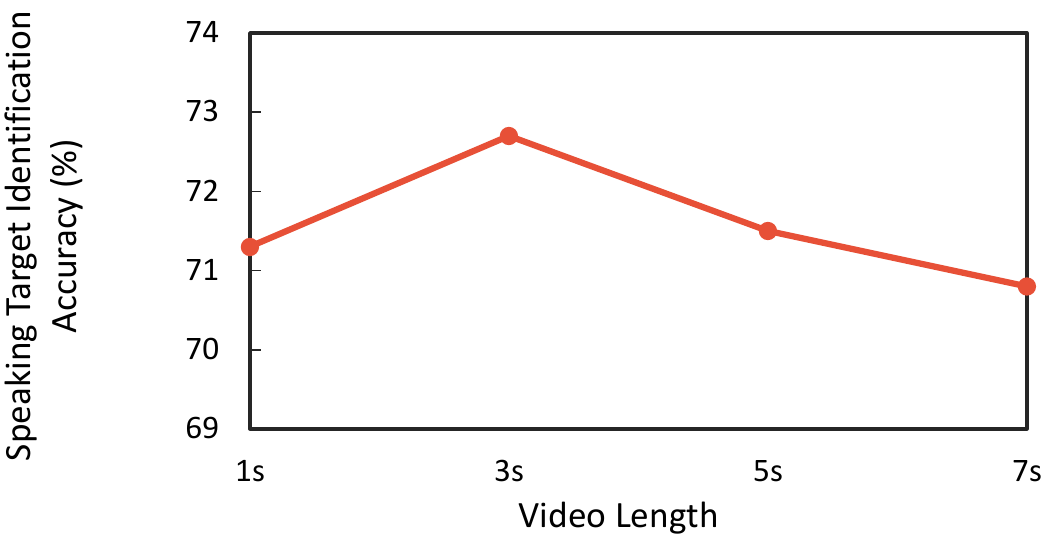}
        \vspace{-0.4cm}
    \end{minipage}
    \begin{minipage}[b]{0.5\textwidth}
    		\renewcommand{\arraystretch}{1.8}
		\renewcommand{\tabcolsep}{5mm}
        \centering
        \resizebox{0.8\linewidth}{!}{%
        \begin{tabular}{c c c}
				\toprule
				 \multirow{1}{*}{\bf Target Task} & {\bf \makecell{Video Length}} &  \bf \multirow{1}{*}{Accuracy (\%)}  \\  \bottomrule 
      \multirow{4}{*}{\makecell{ Speaking Target \\ Identification}} 
    & 1 sec   &  71.3  \\  
    &  \bf \graycell 3 sec   & \bf \graycell  72.7 \\ 
    & 5 sec   &    71.5  \\ 
     &  7 sec   & 70.8  \\  
				\bottomrule
        \end{tabular}
        }
        \vspace{-0.1cm}
    \end{minipage}%
        \captionof{figure}{Effects of video length on the performance for speaking target identification.}
        \label{figure_video_length1}
\end{figure}

\begin{figure}[h]
    \begin{minipage}[b]{0.49\textwidth}
        \centering
        \includegraphics[width=0.8\linewidth]{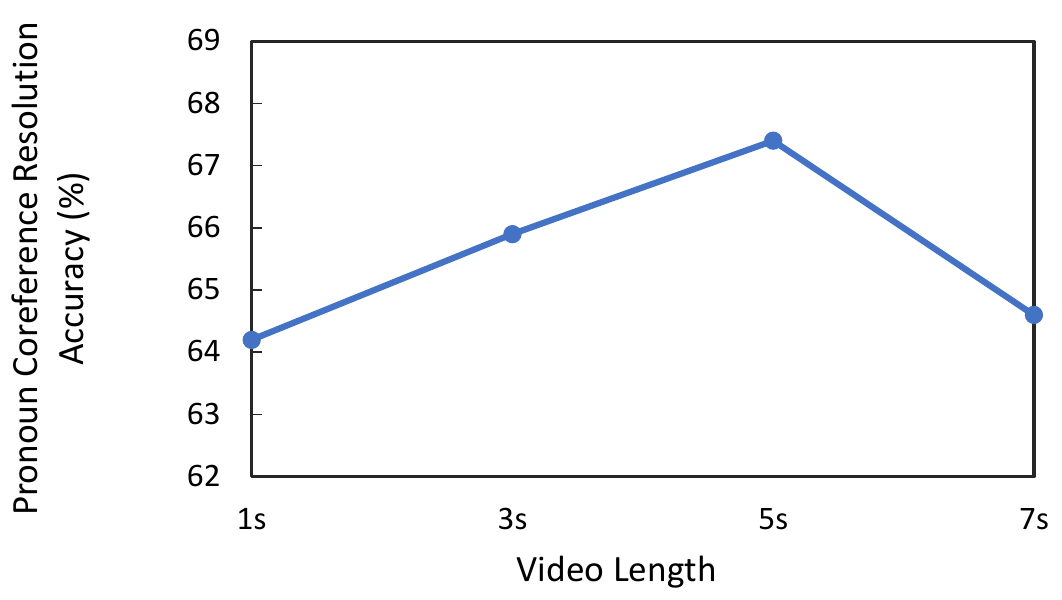}
        \vspace{-0.4cm}
    \end{minipage}
    \begin{minipage}[b]{0.5\textwidth}
    		\renewcommand{\arraystretch}{1.8}
		\renewcommand{\tabcolsep}{5mm}
        \centering
        \resizebox{0.8\linewidth}{!}{%
        \begin{tabular}{c c c}
				\toprule
				 \multirow{1}{*}{\bf Target Task} & {\bf \makecell{Video Length}} &  \bf \multirow{1}{*}{Accuracy (\%)}  \\  \bottomrule 
      \multirow{4}{*}{\makecell{ Pronoun Coreference \\ Resolution}} 
    & 1 sec   &  64.2  \\  
    & \bf \graycell  3 sec   &   \underline{\graycell65.9} \\ 
    & 5 sec   &  \bf  67.4  \\  
     &  7 sec   & 64.6  \\  
				\bottomrule
        \end{tabular}
        }
        \vspace{-0.1cm}
    \end{minipage}%
        \captionof{figure}{Effects of video length on the performance for pronoun coreference resolution.}
        \label{figure_video_length2}
\end{figure}

\begin{figure}[h]
    \begin{minipage}[b]{0.49\textwidth}
        \centering
        \includegraphics[width=0.8\linewidth]{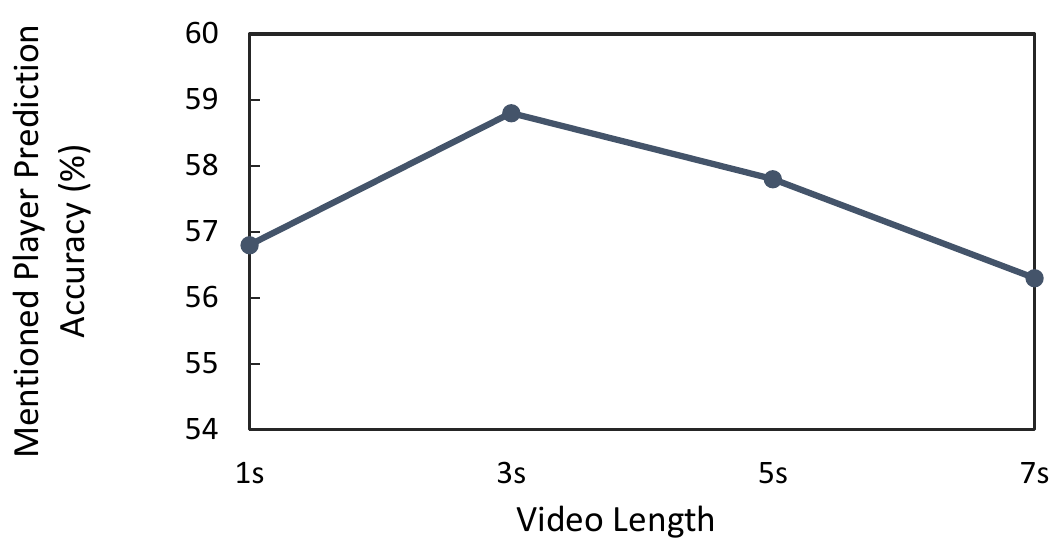}
        \vspace{-0.4cm}
    \end{minipage}
    \begin{minipage}[b]{0.5\textwidth}
    		\renewcommand{\arraystretch}{1.8}
		\renewcommand{\tabcolsep}{5mm}
        \centering
        \resizebox{0.8\linewidth}{!}{%
        \begin{tabular}{c c c}
				\toprule
				 \multirow{1}{*}{\bf Target Task} & {\bf \makecell{Video Length}} &  \bf \multirow{1}{*}{Accuracy (\%)}  \\  \bottomrule 
      \multirow{4}{*}{\makecell{ Mentioned Player \\ Prediction}} 
    & 1 sec   &  56.8  \\  
    &  \bf \graycell  3 sec   & \bf \graycell  58.8 \\ 
    & 5 sec   &  57.8  \\  
     &  7 sec   & 56.3  \\  
				\bottomrule
        \end{tabular}
        }
        \vspace{-0.1cm}
    \end{minipage}%
        \captionof{figure}{Effects of video length on the performance for mentioned player prediction.}
        \label{figure_video_length3}
\end{figure}

We conduct experiments to investigate the effects of video length on the performance of each task. Figures \ref{figure_video_length1}, \ref{figure_video_length2}, and \ref{figure_video_length3} show the validation results for speaking target identification, pronoun coreference resolution, and mentioned player prediction, respectively. We achieve fairly good performance with a video length of 3 seconds for all tasks. Note that we adopt 3 seconds as our default setting for the baselines. These evaluations were conducted on YouTube dataset using the BERT-based model.

\vspace{-0.2cm}
\vspace{-0.8cm}
\vspace{-0.8cm}
\vspace{-0.3cm}
\section{Effects of Player Position Correction}
We address scenarios where a player is temporarily undetected, such as being offscreen for a short time. In such cases, we proceed with player position encoding by leveraging the corresponding player position stored in a buffer to correct the missing player position. Table \ref{table_correction} shows the experimental results demonstrating the impact of missing player correction on the performance of all three social tasks. As shown in the table, the position correction contributes to improved performance. These experiments were conducted using the BERT-based model on YouTube dataset.
\vspace{-1cm}
\begin{table}[h]{
		\renewcommand{\arraystretch}{1.3}
		\renewcommand{\tabcolsep}{5mm}
		\centering
  	\vspace{1.0cm}
		\resizebox{0.6\linewidth}{!}{
			\begin{tabular}{c  c  c }
				\toprule
				 \multirow{1}{*}{\bf Target Task} & {\bf \makecell{Player Position\\ Correction}} &  \bf \multirow{1}{*}{Accuracy (\%)}  \\  \bottomrule 
      \multirow{2}{*}{\makecell{ Speaking Target Identification}} & \xmark   &  70.1   \\  
    & \graycell \cmark   &  \bf \graycell 72.7   \\ \hline

    \multirow{2}{*}{\makecell{ Pronoun Coreference Resolution}} &  \xmark   & 65.3  \\  
    & \graycell \cmark    &  \bf \graycell65.9    \\  \hline

      \multirow{2}{*}{\makecell{ Mentioned Player Prediction}} &  \xmark   & 58.1   \\  
    &  \graycell \cmark    & \bf \graycell58.8   \\  
				\bottomrule
		\end{tabular}}
		
		\captionof{table}{Effects of player position correction on the performances for three social tasks.}
		\label{table_correction}}
		\vspace{-0.5cm}
\end{table}

\section{Data Domain Generalization}
We conduct experiments to validate the generalization capability of our proposed approach across two data domains. To this end, we train our baseline on YouTube dataset and evaluate its performance on Ego4D dataset. Table \ref{table_domain} shows the performance results according to the training data. As shown in the table, the model trained on YouTube data performs well on the Ego4D domain and even achieves better results compared to the model trained on Ego4D for all three social tasks. This improvement can be attributed to the larger amount of training data available in YouTube dataset. The experimental results demonstrate the generalization capability of our approach between different data domains and its potential to work in generalized environments. We adopt the BERT-based model for this experiment.
\vspace{-0.5cm}
\begin{table}[h]{
		\renewcommand{\arraystretch}{1.4}
		\renewcommand{\tabcolsep}{5mm}
		\centering
  	\vspace{1.0cm}
		\resizebox{0.7\linewidth}{!}{
			\begin{tabular}{c  c  c c }
				\toprule
				 \multirow{1}{*}{\bf Target Task}  & {\bf \makecell{Test Data}} & {\bf \makecell{Training Data}} &  \bf \multirow{1}{*}{Accuracy (\%)}  \\  \bottomrule 
      \multirow{2}{*}{\makecell{ Speaking Target Identification}} & \multirow{2}{*}{Ego4D}  & Ego4D  &  61.9   \\  
   &   & \graycell YouTube  &  \bf \graycell 70.5   \\ \hline

    \multirow{2}{*}{\makecell{ Pronoun Coreference Resolution}} & \multirow{2}{*}{Ego4D}  &  Ego4D  & 49.1  \\  
     & & \graycell YouTube  &  \bf \graycell 58.0    \\  \hline

      \multirow{2}{*}{\makecell{ Mentioned Player Prediction}} & \multirow{2}{*}{Ego4D} &  Ego4D   & 50.0   \\  
      & &  \graycell YouTube & \bf \graycell 57.3   \\

				\bottomrule
		\end{tabular}}

		\captionof{table}{Performance results according to the training data types for three social tasks.}
		\label{table_domain}}
	\vspace{1cm}
\end{table}

\vspace{0.5cm}
\section{Additional Quantitative Results}
\noindent \textbf{Utilization of Cropped Visual Features.} We conduct experiments using cropped visual image features for visual interaction modeling. This approach is based on our dense alignment framework but utilizes cropped CLIP \cite{radford2021learning} features instead of keypoint features for the speaker kinesics part (green) in Figure 2 of the main paper. The performances achieved with the cropped CLIP features are 71.0\% for Speaking Target Identification, 63.6\% for Pronoun Coreference Resolution, and 57.7\% for Mentioned Player Prediction. These results are lower compared to our proposed baseline with keypoint features, which achieves 72.7\%, 65.9\%, and 58.8\% for the respective tasks. These evaluations are conducted on YouTube dataset using the BERT-based model.

\noindent \textbf{Measurement of Recall and Precision.} In addition to accuracy, we further measure the macro-precision and macro-recall performance for our proposed approach. It is worth noting that in the multi-class setting, accuracy represents both micro-precision and micro-recall. The results show that our approach achieves (macro-precision / macro-recall / accuracy) performances of (74.8\% / 74.7\% / 72.7\%) for Speaking Target Identification, (64.9\% / 63.3\% / 65.9\%) for Pronoun Coreference Resolution, and (61.9\% / 60.6\% / 58.8\%) for Mentioned Player Prediction. These results are obtained using the BERT-based model on YouTube dataset. We could achieve the balanced performances across precision, recall, and accuracy metrics in our environment, considering precision vs recall and macro vs micro aspects.

\newpage
\section{Additional Qualitative Results}
\begin{figure}[h]
	\vspace{-0.4cm}
	\begin{minipage}[b]{1.0\linewidth}
		\centering
		\centerline{\includegraphics[width=14cm]{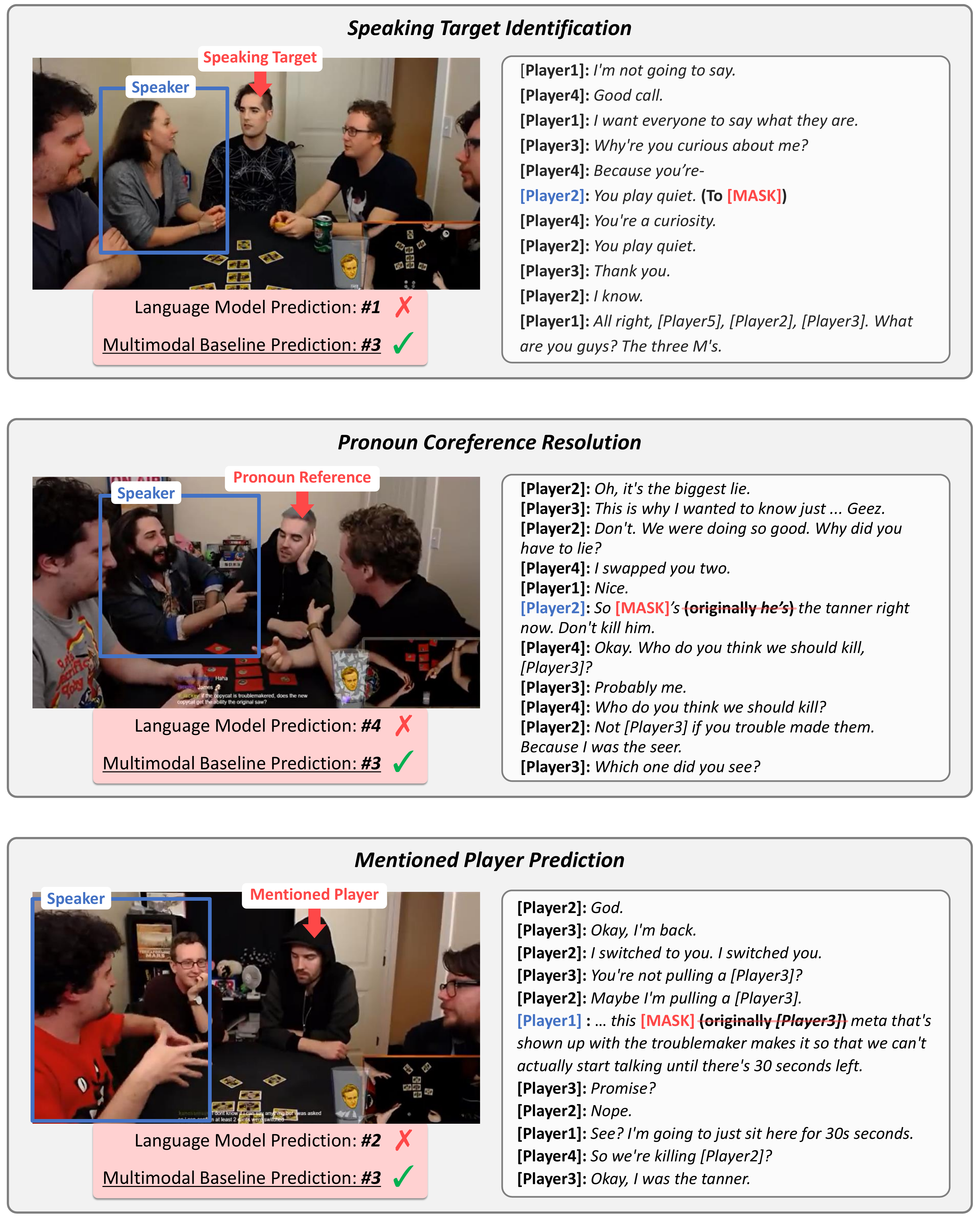}}
	\end{minipage}
\vspace{-0.7cm}
	\caption{Qualitative results demonstrating the benefit of visual cues in multimodal analysis for three social tasks. Note that Player\# are assigned in ascending order from left to right in the visual scenes of this figure.}
	\label{figure_qualitative_supp}
		\vspace{-2.0cm}
\end{figure}

Figure \ref{figure_qualitative_supp} illustrates examples where our multimodal baseline model utilizing aligned language and visual cues outperforms the language-only model across the three social tasks. Our baseline with densely aligned multimodal representations enables corrected inferences compared to relying solely on language input. The BERT model is employed for these experiments. When we make inferences for these samples with RoBERTa language-only model, the inference results are \#3 for STI, \#1 for PCR, and \#2 for MPP, which means it fails on 2nd and 3rd samples. These results demonstrate cases where the multimodal model (\textit{i.e.}, BERT) with visual reasoning surpasses the more powerful language-only model (\textit{i.e.}, RoBERTa).
\vspace{0.2cm}

\newpage
\section{Social Deduction Game Details}
We leverage two social deduction game datasets: YouTube and Ego4D. Note that the data collections and annotations have been approved by the Institutional Review Board (IRB). YouTube dataset includes the games of \textit{One Night Ultimate Werewolf} while Ego4D dataset contains the games of \textit{One Night Ultimate Werewolf} and \textit{The Resistance: Avalon}. Below are the details for each social deduction game: Werewolf and Avalon.

\noindent \textbf{One Night Ultimate Werewolf.} One Night Werewolf is a social deduction game in which players are secretly assigned to one of two primary factions - the villager team or the werewolf team. During the night phase, players close their eyes and characters with special abilities perform actions like swapping cards before opening their eyes again. The night phase may alter players' roles, though most remain unaware of these changes. Subsequently, players engage in discussion and negotiation to deduce the werewolf's identity. Werewolves, on their part, strive to conceal their identity and mislead others. At the end, everyone votes on who they believe is most suspicious. If at least one werewolf is eliminated, the village team wins, but if no werewolves are eliminated, the werewolf team wins. We refer One Night Ultimate Werewolf game's rules on Wikipedia \url{https://en.wikipedia.org/wiki/Ultimate_Werewolf}.

\noindent \textbf{The Resistance: Avalon.} This game splits players into two groups: the Minions and the Loyal Servants of Arthur. After roles are assigned secretly via card distribution, players commence with a round where each assumes the Leader role in turns. The Leader's role involves proposing a team for a Quest and all players discuss and vote on approving or rejecting the team assignment. Post the Team Building phase, the designated team decides the Quest's outcome. The Good Team is restricted to using only the Success card in the Quest phase, whereas the Evil Team has the option to use either the Success or Fail card. The Good Team claims victory upon completing three successful Quests, while the Evil Team wins either by causing three Quests to fail or by correctly identifying the character Merlin among the Good Team. We refer The Resistance: Avalon game's rules on Wikipedia \url{https://en.wikipedia.org/wiki/The_Resistance_(game)}.



\end{document}